# Word Alignment-Based Evaluation of Uniform Meaning Representations

**Daniel Zeman, Federica Gamba**
Charles University
Faculty of Mathematics and Physics
Institute of Formal and Applied Linguistics (ÚFAL)
Prague, Czechia
{zeman, gamba}@ufal.mff.cuni.cz

**Abstract**

Comparison and evaluation of graph-based representations of sentence meaning is a challenge because competing representations of the same sentence may have different number of nodes, and it is not obvious which nodes should be compared to each other. Existing approaches favor node mapping that maximizes $F_1$ score over node relations and attributes, regardless whether the similarity is intentional or accidental; consequently, the identified mismatches in values of node attributes are not useful for any detailed error analysis. We propose a node-matching algorithm that allows comparison of multiple Uniform Meaning Representations (UMR) of one sentence and that takes advantage of node-word alignments, inherently available in UMR. We compare it with previously used approaches, in particular *smatch* (the de-facto standard in AMR evaluation), and argue that sensitivity to word alignment makes the comparison of meaning representations more intuitive and interpretable, while avoiding the NP-hard search problem inherent in *smatch*. A script implementing the method is freely available.

## 1. Introduction

In many frameworks, meaning of natural language is represented as a directed graph structure (e.g., Žolkovskij and Mel'čuk (1967); Sgall et al. (1986); Ivanova et al. (2012); Abend and Rappoport (2013); Banarescu et al. (2013); van Gysel et al. (2021); see Žabokrtský et al. (2020) for a survey). Nodes of the graph typically represent "word-like" units; nevertheless, 1-1 correspondence between nodes and surface words is not guaranteed, and in some frameworks, no node-word alignment is defined. Consequently, such structures are harder to produce, both by human annotators and by parsing models. It is also not easy to evaluate the parser output against gold standard graphs, or to compare graphs produced by two independent annotators.

In this paper, we discuss evaluation of Uniform Meaning Representation (UMR) (van Gysel et al., 2021; Bonn et al., 2024), a typologically aware extension of Abstract Meaning Representation (AMR) (Banarescu et al., 2013). In both AMR and UMR, a sentence is represented by a directed graph of entities and events (Figure 1). Each node has a unique identifier, e.g., `s23w`, and a concept string, e.g., `want-01`.[1] Relations between nodes are labeled, e.g., the relation from an event to a time expression can be labeled `:temporal`. Nodes can also have attributes, which are similar to relations, but instead to a child node they point to the value of the attribute (e.g., `:refer-number plural`). In addition, UMR introduces document-level annotation of temporal relations, coreference and modality; these are labeled relations between nodes that may occur in different sentences. Finally, UMR files[2] also explicitly show the surface tokens and their alignment with concept nodes. This contrasts with AMR, which is by design agnostic to how strings are related to meanings.

[1]An alternative view in the literature has the identifiers called *variables*, they refer to inner nodes of the graph, and concepts are shown as separate leaf nodes, connected to their variables as their *instances*.

## 2. The Problem

Comparison of semantic graphs is a non-trivial task because two representations of the same sentence may differ in the number of nodes, and the node identifiers typically differ, too. It is thus not obvious which nodes should be taken as corresponding to each other. If we can find the optimal node mapping between the two graphs, the rest of the task is easy. Properties of the graph can be expressed as a set of triples $(x, y, z)$, where $x$ is a node (now identifiable in both graphs), $y$ is a name of a relation or an attribute, and $z$ is another node (child node of the relation) or the value of the attribute. Furthermore, the node's concept can be seen as a special kind of attribute, yielding the triple $(x, \texttt{:concept}, \textit{concept})$. Similarity of two graphs can be expressed as the $F_1$ score of the triples.

[2]At the time of writing, UMR release 2.0 is available at http://hdl.handle.net/11234/1-5902 and it contains datasets of 8 languages.

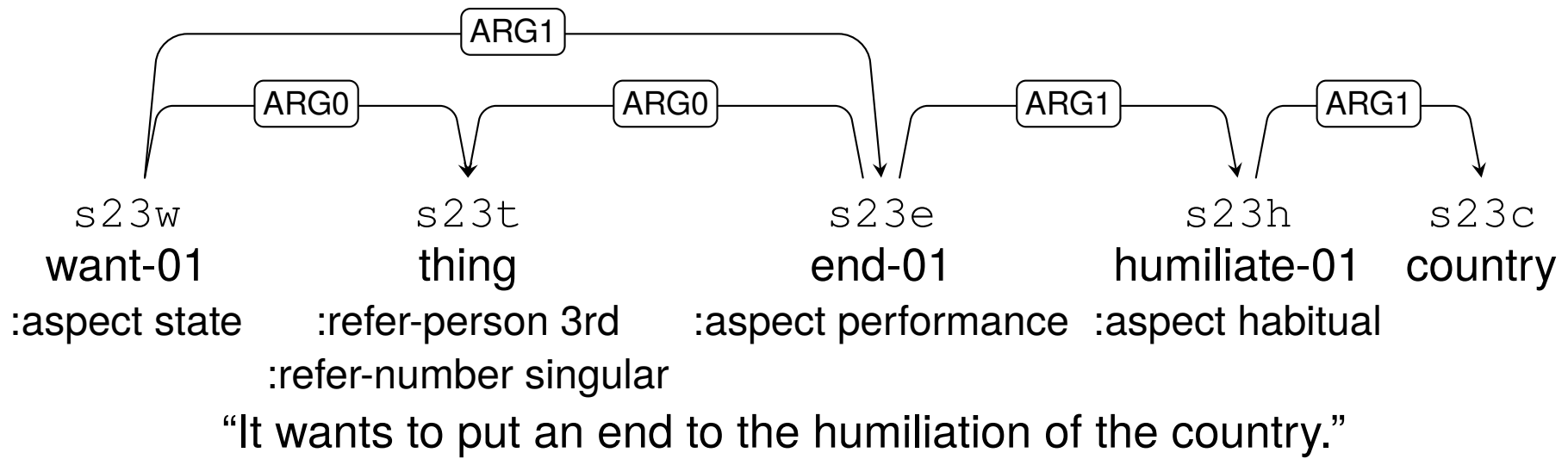

Figure 1: Example of a sentence-level graph in UMR, taken from the English data in UMR 2.0. Concepts are shown below node ids, subsequent lines show node attributes and their values.

Note that if we can map concept nodes between two UMRs, we can also evaluate the document-level relations (modal, temporal, and coreferential), as they can be expressed as triples using the same node mapping, regardless of the fact that they connect nodes in different sentences.

## 3. Related Work

For AMR, the *smatch* metric (Cai and Knight, 2013) has emerged as the de-facto standard. It defines as optimal the mapping that maximizes $F_1$ of the resulting triples; the *smatch* algorithm employs hill-climbing with restarts to find an approximate solution to the NP-hard optimization problem.

A more general evaluation algorithm, *mtool*, was used in the Meaning Representation Parsing shared tasks at CoNLL (Oepen et al., 2020). It can evaluate graph structures defined in various frameworks, including AMR. Instead of greedy hill-climbing, it uses the MCES algorithm by McGregor (1982) with pre-set search space limits. Word alignment (called *anchoring* in the task) has no special role in node matching, but it can be evaluated together with other node attributes.

An alternative node mapping algorithm, called *AnCast*, has been proposed specifically for UMR (Sun and Xue, 2024; Sun et al., 2025). It has been shown to be more efficient and more accurate than *smatch*. The authors also define partial metrics such as Concept $F_1$ and Labeled Relation $F_1$, which improve interpretability of the results.

*Smatch* does not have the notion of word alignment; *AnCast* can use it if available, but it can work without it, too. Nevertheless, *AnCast's* ability to exploit alignment is limited. The token–node alignments can be $M : N$, with a node potentially mapped to a discontinuous set of tokens, while *AnCast* can currently process only continuous alignments. *AnCast* also compares concepts of the nodes to be mapped, and it tries to assess concept similarity rather than identity, although in a restricted manner. To achieve similarity $> 0$, one concept lemma must be substring of the other. This would recognize similarity between e.g., *fry* and *stir-fry*, but not between Czech *volit* 'to vote' and its nominalization *volba* 'election'.

## 4. Yet Another Approach: ju:mætʃ

Our goal is twofold: Besides numeric evaluation of graph similarity, we also want to obtain node mapping to highlight and eyeball disagreement between annotators (or between system output and gold data). We argue that the existing mapping approaches are not well suited for the latter purpose.

Both *smatch* and *AnCast* will map as many nodes as possible. If one of the graphs has more nodes than the other, remaining nodes will stay unmapped. If the graphs have the same number of nodes, every node will be mapped to a node in the other graph, even if they are clearly unrelated. This may occasionally improve the score when a random attribute occurs in both nodes, but it blurs the interpretation of the score, and any (dis)agreement in attribute values of such nodes is meaningless. Consider the two graphs in Figure 2. Shared attributes and one relation will lead to mapping `xh-yi`, `xp1-yp2`. Identical concept `business` will add `xb-yb`. Likewise, the concept `person` will grant the mapping `xp2-yp1`, but these two persons should not be equated (`xp2` is the role for 'you', while `yp1` is the speaker). Finally, `xm` will not be identified with `ym`: despite both of them being motivated by the word 'manager', their concepts differ. Instead, the shared aspect feature will drag together semantically unrelated `xm-ys`.

Therefore, we propose another mapping algorithm called *ju:mætʃ*,[3] which primarily maps nodes aligned to the same word(s), and for nodes without word alignment (assumed to be a minority in UMR graphs) it requires concept identity. As with

[3]Phonetic for u-match, i.e., "matching UMR."

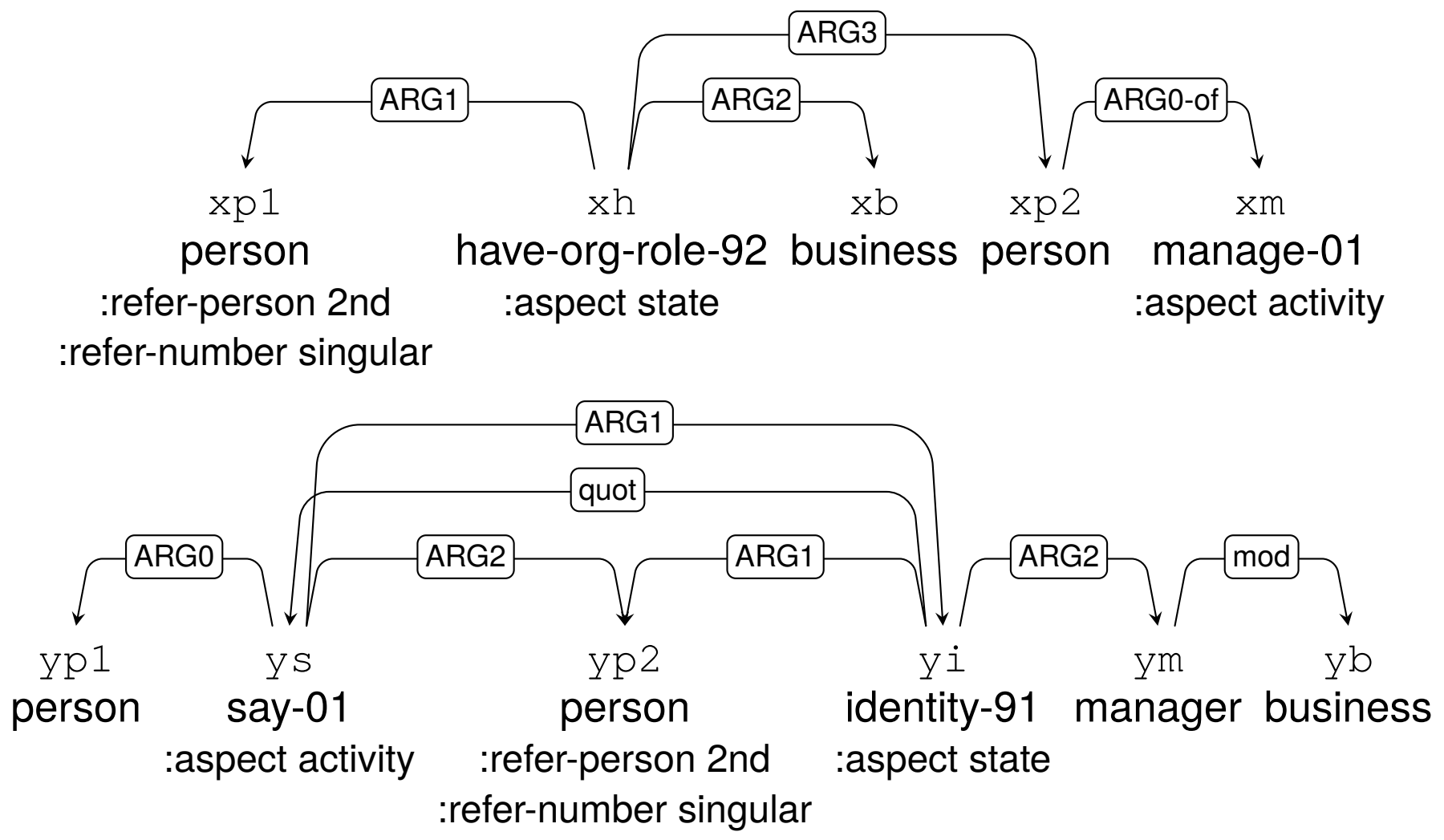

Figure 2: Hypothetical competing annotations of "You are the manager of the business."

*smatch* and *AnCast*, we assess similarity of other attributes if needed to symmetrize the mapping.

As word alignment is part of the annotation, the metric evaluates it, too, though only indirectly. Missing or nonsensical alignments will lead to suboptimal node mapping and thus lower scores.[4] On the other hand, using imperfect alignment for node matching is less straightforward than it may seem. A node may be aligned to multiple words; alignments of nodes from two annotations of the same sentence may overlap instead of being identical. For example, in Figure 1, node `s23w` may be aligned to *wants* and `s23e` to *put an end* (but one could also decide to include the infinitive marker *to* in the alignment). A different UMR over the same sentence might align one node to *wants to put* and another to *an end*. Overlapping alignments have to be resolved so that at most one node from either side is retained. We need symmetric one-to-one node mappings—not only because it simplifies subsequent comparison of node properties, but also because it follows the intuition that both nodes were intended to represent the same concept in the meaning structure of the sentence.

The symmetrization works as follows. Whenever a node on either side is mapped to multiple nodes on the other side, mappings are gradually removed until just one target node remains. When removing target nodes, we try to remove those that are least similar to the source node, according to several criteria. The most important criterion is concept string identity. If it does not lead to unique node mapping, we also consider all attributes of the node and their values (outgoing relations are treated analogously). Next, we also do a "weak" comparison of attributes, where we only consider the presence of an attribute, without requiring identical value of the attribute on both sides. This can help distinguish e.g. eventive concepts (having attributes such as `:aspect`) from entities (having e.g. `:refer-number` or `:name`). Finally, we also prioritize alignment to longer words (trying to avoid relying too much on function words, which tend to be shorter).

Nodes without word alignment are paired when they have the same concept; if there are multiple options, we use the same symmetrization approach as described above. The assumption is that many of them are abstract UMR concepts such as `identity-91`, `have-mod-91` or `name`, hence comparing their concept strings will often point at the correct mapping.[5]

Unlike *smatch*, the *ju:mætʃ* approach is not NP-hard.[6] Like *AnCast*, we compute partial metrics that help further pinpoint issues with the annotation, such as separate $F_1$ of concepts, labeled relations or selected node attributes. Moreover, we can deal with discontinuous word alignment. An implementation of *ju:mætʃ* graph comparison is available and open-source.[7]

[4] A more direct evaluation of alignment is implemented as a secondary metric – $F_1$ of subsets of tokens that are aligned to one node in each graph.

[5] Specifically for the `name` concept, we enhance it with the real name from its `:opN` attributes before comparing it to other nodes, e.g., `name["United" "States"]`. This helps in sentences with multiple named entities.

[6] The symmetrization makes our approach potentially $O(n^2)$, although in practice $M : N$ mappings are not frequent: In our test runs, they affected about 6% of all nodes, and the average ambiguity was $1:2$.

[7] https://github.com/ufal/umrtools

# 5. Evaluation

It is not possible to do a large-scale test of the proposed method – datasets with double annotations are not publicly available, and the research on UMR parsers has also been quite limited so far. The minimum we can do is to illustrate the metric on a simple baseline parser. We took the first 5 English files (209 sentences) from the UMR 2.0 release as the gold standard, and compared them with the same sentences automatically parsed to Universal Dependencies[8] and then converted to UMR.[9] Table 1 shows the scores contrasted with *smatch* and *AnCast*. The numbers are low because a substantial share of necessary information is missing in the syntactic dependencies; however, our focus here is not on the conversion method but on the way how converted UMR graphs are mapped on those from the gold standard. The scores presented here do not include document-level relations because the UD-to-UMR conversion does not produce them; also, the existing implementation of *smatch*[10] expects only sentence graphs on input (although in principle, *smatch* could be extended to document-level triples as well).

| | **P** | **R** | **$F_1$** |
|---|---|---|---|
| *smatch* | 32.89 | 31.89 | 32.39 |
| *AnCast* | 18.36 | 18.56 | 18.46 |
| *ju:mæʧ* | 25.44 | 24.66 | 25.05 |

Table 1: The three metrics contrasted on the same set of English UMR files (converted dependency trees vs. gold standard annotation).

The sample contains 1675 gold nodes and 1501 converted nodes, respectively (in both cases, there are over 4500 triples to compare). *Ju:mæʧ* mapped 1157 nodes from each side, hence mapping precision $P = 77\%$ and recall $R = 69\%$.[11] In contrast, *smatch* will match as many nodes as possible, leading to $P = 100\%$ and $R = 90\%$. If we evaluate only the triples on the nodes successfully mapped, the *ju:mæʧ* scores will approach those of *smatch*: $P = 32\%$, $R = 34\%$, and $F_1 = 33\%$.

The scores themselves cannot verify the assumption that leaving some nodes unmapped is a good idea. To shed some light on that, a single annotator (one of the authors) examined the first of the 5 English documents, answering the question whether one of the unmapped nodes should actually be mapped to another node, which was also left unmapped. There were 203 unmapped nodes in the sample (123 in gold and 80 in converted UMRs). Out of these, 166 (82%) were found correct and 37 (18%) wrong.

| **Language** | **Nodes** | **Aligned** | **Unaligned** |
|---|---|---|---|
| Czech | 2,094,192 | 90% | 10% |
| Arapaho | 537 | 85% | 15% |
| Latin | 773 | 82% | 18% |
| English | 4,178 | 80% | 20% |
| Chinese | 27,736 | 64% | 36% |
| Navajo | 3,255 | 61% | 39% |
| Kukama | 525 | 55% | 45% |
| Sanapana | 2,338 | 47% | 53% |

Table 2: The proportion of word-aligned and unaligned nodes in UMR 2.0.

As for our assumption that nodes without word alignment are relatively rare, it does not necessarily hold (or not to the same extent) for all languages and domains. In the English sample evaluated above, the converted file had only 11% unaligned nodes, but the gold standard had 24%. As seen in Table 2, the percentage of unaligned nodes in UMR 2.0 ranges from 10% (Czech) to 53% (Sanapana), although the overall proportion is 11% due to the very large size of the Czech dataset. While in general a higher number of unaligned nodes could be caused by the annotator's/parser's failure to annotate all alignments, the particular case of Sanapana in UMR 2.0 seems to be justified. There is a lot of unexpressed implicit subjects, which are represented by abstract nodes. In addition, abstract predicates are quite frequent, too, as reported by Bonn et al. (2024): "[...] absence of reporting verbs in narrations, with speakers simply switching back and forth [...] This required the annotation of many 'say'-events [...] that were not explicitly present in the Sanapaná text." Therefore, the combination of language features (pro-drop), and text type (1st person narration) may lead to structures where our emphasis on word alignment is less advantageous.

# 6. Conclusion

We presented *ju:mæʧ*, a node matching algorithm for Uniform Meaning Representations, which takes advantage of node-word alignments. We argue that it is particularly useful for error analysis when individual differences between representations are to be inspected. At the same time, node matching leads straightforwardly to $F_1$ score that reflects the quality of conversion or parser output, or inter-annotator agreement. Our implementation also computes detailed partial scores that allow assessing specific issues, such as differences in individual attributes or relations. The tool is open-source and has been used in annotation of Czech and Latin UMR.

[8] The parses were obtained using UDPipe v2 (Straka, 2018) with the English-GUM v2.15 pretrained model.

[9] In principle, conversion to UMR from another annotation framework can be compared with gold standard annotation (i.e., manual annotation directly in UMR) just like the output of a UMR parser. Conversion efforts have been reported from English AMR (Bonn et al., 2023), from Czech and Latin tectogrammatical structures (Štěpánek et al., 2025), and from UD trees in many languages (Gamba et al., 2025).

[10] https://github.com/snowblink14/smatch

[11] Note that here $P$ and $R$ only reflect how many nodes were mapped; not the "correctness" of the mapping.

## 7. Limitations

The focus on node-word alignment, which we consider the biggest advantage of our method, is at the same time a limitation. At present, there are no defined guidelines for word-node alignment in UMR, and different annotation projects, annotators, and languages may adopt varying alignment strategies. This variability introduces potential inconsistencies that can affect both evaluation and cross-linguistic comparison. Additionally, the use of concept identity as the second mapping criterion may lead to unintended outcomes in cases where there are significant errors or discrepancies in concept assignment.

## 8. Acknowledgments

The work described herein has been supported by the grants *LINDAT/CLARIAH-CZ* (Project No. LM2023062) of the Ministry of Education, Youth, and Sports of the Czech Republic, and GAUK No. 104924 of the Charles University.

The project has been using data and tools provided by the *LINDAT/CLARIAH-CZ Research Infrastructure* (`https://lindat.cz`), supported by the Ministry of Education, Youth and Sports of the Czech Republic (Project No. LM2023062).

## 9. Bibliographical References

## A. Node mapping in *juːmætʃ* and *smatch*

Here we show an example sentence from gold and converted Czech UMR data and document the word alignments and the node mapping used by the two metrics.

The full sentence: *Vážení čtenáři, je tomu právě rok, kdy jsme vám oznamovali nepopulární informaci, že se cena našich novin zvyšuje.* "Dear readers, it's been a year since we announced the unpopular news that the price of our newspaper was increasing."

Our excerpt: *Vážení čtenáři, je tomu právě rok, kdy jsme vám oznamovali informaci* "Dear readers, it's been a year since we announced the news"

```
GOLD:
(s1p0 / publication-91
  :ARG3 (s1s1 / say-91
    :aspect activity
    :modal-strength full-affirmative
    :ARG0 (s1p1 / person
      :refer-number plural
      :refer-person 1st)
    :ARG2 (s1p2 / person
      :refer-number plural
      :refer-person 2nd
      :ARG0-of (s1c1 / číst-002 'read'
        :aspect habitual
        :modal-strength full-affirmative)
      :mod (s1v1 / vážený 'dear'))
  :ARG1 (s1h1 / have-temporal-91
    :aspect state
    :modal-strength full-affirmative
    :quote s1s1
    :vocative s1p2
    :ARG1 (s1o1 / oznamovat-002 'announce'
      :aspect performance
      :modal-strength full-affirmative
      :ARG0 s1p1
      :ARG1 (s1i1 / informace 'information'
        :refer-number singular)
      :ARG2 s1p2)
    :ARG2 (s1r1 / rok 'year'
      :refer-number singular
      :mod (s1p4 / právě 'just')))))

CONV:
(s1b1 / být-011
  :aspect activity
  :vocative (s1c1 / čtenář 'reader'
    :refer-number plural
    :mod (s1v1 / vážený 'dear'))
  :ARG1 (s1t1 / ten
    :refer-number singular
    :temporal (s1p1 / právě 'just'))
  :duration (s1r1 / rok 'year'
    :refer-number singular
    :temporal-of (s1o1 / oznamovat-002 'announce'
      :aspect activity
      :ARG0 (s1p2 / person
        :refer-number plural
        :refer-person 1st)
      :ARG1 (s1i1 / informace 'information'
        :refer-number singular)
      :ARG2 s1c1)))
```

*juːmætʃ* node mapping between GOLD and CONV (word alignment, if any, is shown in brackets after the concept):

```
s1p0 / publication-91 … UNMAPPED
s1s1 / say-91 … UNMAPPED
s1p1 / person ("našich") … s1p2 / person ("našich")
s1p2 / person ("čtenáři vám")
       … s1c1 / čtenář ("čtenáři vám")
s1c1 / číst-002 … UNMAPPED
s1v1 / vážený ("Vážení") … s1v1 / vážený ("Vážení")
s1h1 / have-temporal-91 ("je") … s1b1 / být-011 ("je")
s1o1 / oznamovat-002 ("tomu jsme oznamovali")
       … s1o1 / oznamovat-002 ("jsme oznamovali")
s1i1 / informace ("informaci")
       … s1i1 / informace ("informaci")
UNMAPPED … s1t1 / ten ("tomu")
s1r1 / rok ("rok kdy") … s1r1 / rok ("rok kdy")
s1p4 / právě ("právě") … s1p1 / právě ("právě")
```

*smatch* node mapping between GOLD and CONV (showing only differences from *juːmætʃ* mapping):

```
s1s1 / say-91 … s1b1 / být-011 ("je")
s1h1 / have-temporal-91 ("je") … s1t1 / ten ("tomu")
```

In our excerpt, the only nodes left unmapped by *smatch* are `s1p0` and `s1c1` from the GOLD graph, because there are no nodes left available in the CONV graph. There are two other nodes that are left unmapped by *juːmætʃ* but not by *smatch*: `s1s1` in GOLD and `s1t1` in CONV. The mapping that *smatch* found for these nodes has no semantic justification (but it will slightly increase $F_1$ score because both `say-91` and `být-011` have `:aspect activity`).